# Designing Intelligent Instruments

Kevin H. Knuth[a,b], Philip M. Erner[a], Scott Frasso[c]

a. Univ. at Albany, Dept of Physics, Albany NY USA
b. Univ. at Albany, Dept of Informatics, Albany NY USA
c. Northeastern Univ., Dept. of Electrical and Computer Engineering, Boston MA USA
http://knuthlab.rit.albany.edu/

**Abstract.** Remote science operations require automated systems that can both act and react with minimal human intervention. One such vision is that of an intelligent instrument that collects data in an automated fashion, and based on what it learns, decides which new measurements to take. This innovation implements experimental design and unites it with data analysis in such a way that it completes the cycle of learning. This cycle is the basis of the Scientific Method.

The three basic steps of this cycle are hypothesis generation, inquiry, and inference. Hypothesis generation is implemented by artificially supplying the instrument with a parameterized set of possible hypotheses that might be used to describe the physical system. The act of inquiry is handled by an inquiry engine that relies on Bayesian adaptive exploration where the optimal experiment is chosen as the one which maximizes the expected information gain. The inference engine is implemented using the nested sampling algorithm, which provides the inquiry engine with a set of posterior samples from which the expected information gain can be estimated. With these computational structures in place, the instrument will refine its hypotheses, and repeat the learning cycle by taking measurements until the system under study is described within a pre-specified tolerance. We will demonstrate our first attempts toward achieving this goal with an intelligent instrument constructed using the LEGO MINDSTORMS NXT robotics platform.

**Keywords:** intelligent, robotics, experimental design, automation, instrumentation
**PACS:** 07.05.Bx, 07.05.Dz, 07.05.Fb, 07.05.Hd, 07.05.Kf

## INTRODUCTION

Remote science operations are currently being carried out using robotic explorers both on Mars and in deep sea studies here on Earth. These operations, which employ semi-automated systems that can carry out basic tasks such as locomotion and directed data collection, require human intervention when it comes to deciding where to go, which experiment to perform, and precisely where to place the sensors. However, as we expand to explore more remote worlds, we will require that our instruments be increasingly autonomous. The vision we present in this paper is that of an intelligent instrument that collects data in an automated fashion, and based on what it learns, the instrument decides which new measurements to take. The innovation we describe implements automated experimental design and unites the process with automated data analysis in such a way that it completes the cycle of learning.

Many researchers have worked on the problem of designing intelligent systems. Relevant to our approach are the concepts of cybernetics (Wiener, 1948) and experimental design (Lindley, 1956; Fedorov, 1972), which have been pursued and unified in various forms by several researchers. Of particular note is the work on cybernetics by Fry (2002), the active data selection approach of MacKay (1992), and maximum entropy sampling and Bayesian experimental design by Sebastiani and Wynn (2000). The maximum entropy sampling approach to experimental design was expounded upon by Loredo (2003) in his work on Bayesian adaptive exploration, which forms the basis of the approach we present here.

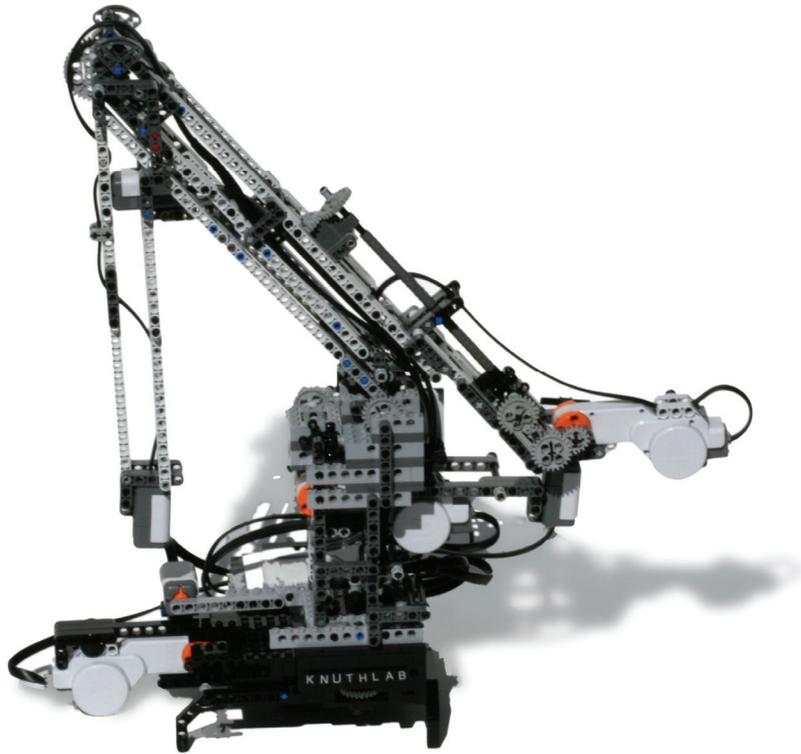

**FIGURE 1.** A photograph of the robotic arm. The end of the arm is equipped with a light sensor that can make point measurements. The robot is built using the LEGO MINDSTORMS NXT system, and is locally controlled with the NXT brick. The NXT brick can communicate with a laptop computer using Bluetooth. The laptop computer (not shown) runs the inference and inquiry code in MATLAB. At the time of the workshop, the MATLAB to NXT communication was not completely operable, and the system was demonstrated via simulations.

The first author of this paper, having been inspired by the work of Cox (1979) and Fry (2002), has been actively developing a calculus for questions (Knuth, 2002, 2003, 2005, 2006) based on bi-valuations on lattices (Knuth, 2007) with an explicit focus on experimental design. However, this framework, which is still in its infancy, is not yet suited for our efforts here. Instead, we employ proven computational technologies.

To create an intelligent instrument, we require three steps: hypothesis generation, experimental design, and data analysis. Hypothesis generation is implemented by programming the instrument with a parameterized model that represents a set of hypotheses that could be used to describe the physical system. Experimental design, which is an act of inquiry, is implemented using Bayesian adaptive exploration (Loredo, 2003), where the optimal experiment maximizes the expected information gain. Finally the data analysis, or inference, is handled using nested sampling

(Skilling, 2005; Sivia & Skilling, 2006), which allows us to test various hypotheses given the newly collected data. At each stage, the instrument will refine its hypotheses and repeat the cycle taking measurements until the system is described within a pre-specified tolerance. In the following sections, we describe our work in the context of a robotic arm solving a characterization problem.

## THE EXPERIMENTAL SETUP

Choosing a problem that is at the same time interesting, challenging, and enlightening is extremely difficult. The problem we have chosen is indeed a toy problem, but one that is easily extended to problems encountered in the real world. We consider an instrument that is designed to locate and characterize a white circle on a black field.

### The Experimental Problem

We have developed a robotic instrument that is designed to locate and characterize a white circle on a black background. The instrument is equipped with a light sensor which is able to take point measurements. We have purposely designed the system so that the sensor cannot simply scan the visual scene. Such scans result in numerous non-informative measurements that waste time, energy and transmission bandwidth. This limited sensor capability is intentional and will serve to highlight the power of the computational techniques we are developing. In addition, the light sensor has a rather large point spread function, which we will not consider in this initial presentation. Instead, we assume that the light sensor returns a measurement that is normally distributed about the mean light intensity, and ignore "edge-effects".

This is clearly a search problem using an instrument with limited sensor capability. As such, the results here are readily extended to similar problems, such as land mine detection. To characterize the circle, the instrument will continue to take measurements until both the center position of the circle and its radius are known to within a predefined accuracy. Those familiar with information theory will realize that once the white circle has been detected, on average, only a small number of binary questions will be necessary to achieve this. We will show that our results agree with this expectation.

### The Robot and its Brains

The instrument is a robotic arm built with the LEGO MINDSTORMS NXT system (Figure 1). The arm has three degrees of freedom, with the ability to rotate about the vertical axis (z-axis), and at two points about the y-axis (elbow and wrist). This gives the arm access to a large region of the horizontal plane. The light sensor, which is mounted at the end of the arm, is constrained to point vertically downward at all times.

The LEGO MINDSTORMS NXT Brick is the computer that directly controls the motors and sensors of the robot. The Brick is programmed in the NXT-G programming language, which is a variant of LabVIEW. The Brick has been programmed with a simple program that moves the arm from the home position to a position on the plane and records the light intensity. After writing the measurement result to a file, the arm returns to the home position.

The intelligence of the robot lives on a Dell Latitude D610 laptop computer. The software is programmed in MATLAB and operates within the Windows XP operating system. The laptop computer communicates with the robot via a Bluetooth Wireless connection to the LEGO Brick. The MATLAB software interacts with the Brick by reading files, writing files and starting programs on the Brick. To request a measurement at a specified location, the MATLAB software must compute the number of motor rotations for each motor and write these values to a file on the LEGO Brick. MATLAB then starts the motor program on the Brick, which reads this file and implements the instructions. When the robot is finished it creates a file containing the resulting light level value. The MATLAB software then reads this file to obtain the data and begin its analysis and evaluation.

While both the MATLAB and the Brick software are operational, we were unable to implement the MATLAB to NXT communication by the time of the workshop. Instead, our experiments were performed with the files being transferred manually.

## INFERENCE AND INQUIRY

To accomplish this task in an intelligent manner, the instrument must be endowed with both an inference engine and an inquiry engine. The inference engine relies on Bayesian methods to infer the circle parameters from the acquired data. The inquiry engine relies on the posterior density over the space of circles to evaluate which measurement is expected to deliver the greatest amount of information. The following subsections describe these two engines.

### The Inference Engine

We begin with the problem of using the available data to infer the circle parameters

$$\mathbf{C} = \{(x_o, y_o), r\} \quad (1)$$

where $(x_o, y_o)$ is the circle center coordinates, and $r$ is the circle radius. In this case, the data consist of a set of $N$ light measurements taken at various points on a plane. We will denote these measurements collectively as $\mathbf{D}$, and write them individually as

$$\mathbf{D} = \{d_1, d_2, \ldots, d_N\} \quad (2)$$

recorded at positions

$$\mathbf{X} = \{(x_1, y_1), (x_2, y_2), \ldots, (x_N, y_N)\}. \quad (3)$$

In this initial exploration, the positions are assumed to be known with certainty.

The goal is to explore the posterior probability

$$p(\mathbf{C} \mid \mathbf{D}, \mathbf{X}, I) = p(\mathbf{C} \mid I) \frac{p(\mathbf{D} \mid \mathbf{C}, \mathbf{X}, I)}{p(\mathbf{D} \mid I)}, \quad (4)$$

where *I* represents our prior information. From this we can obtain a set of posterior samples, each representing a possible circle. We accomplish this using the nested sampling algorithm, which samples from the prior probability and explores within an ever-contracting hard likelihood constraint (Skilling, 2005; Sivia & Skilling, 2006). There are multiple benefits to this approach. First, the algorithm provides a set of posterior samples, which are later used by the inquiry engine to select measurement locations. Second, nested sampling produces an estimate of the evidence, which can be used in the event that the robot needs to test one model against another. A simple example of this would be if the robot is designed to identify whether the white object is a circle or a square. However, in this initial exploration, we focus only on circles.

Here we keep the probability assignments as simple as possible and assign uniform distributions over reasonable ranges of values

$$p(x_o \mid I) = (x_{max} - x_{min})^{-1} \tag{5}$$

$$p(y_o \mid I) = (y_{max} - y_{min})^{-1} \tag{6}$$

$$p(r \mid I) = (r_{max} - r_{min})^{-1}. \tag{7}$$

The results we present here are based on simulations on a playing field of 20cm x 30cm, so that $r_{min} = 1$cm and $r_{max} = 15$cm. By assigning the prior for the center position to be independent of the prior for the radius, we are stating that the entire circle may not be in the playing field. This poses no problem for this investigation.

The likelihood function is again greatly simplified for these simulations. We do not consider the point-spread function of the light sensor and instead assume that the sensor will record the light intensity directly below the sensor with some Gaussian noise. The likelihood for one measurement $d_i$ taken at $(x_i, y_i)$ can be written as

$$p(d_i \mid \mathbf{C}, (x_i, y_i), I) \equiv p(d_i \mid \{(x_o, y_o), r\}, (x_i, y_i), I). \tag{8}$$

$$= \begin{cases} N(d_W, \sigma) & \text{if } (x_i - x_o)^2 + (y_i - y_o)^2 \leq r^2 \\ N(d_B, \sigma) & \text{if } (x_i - x_o)^2 + (y_i - y_o)^2 > r^2 \end{cases}$$

where $N(\cdot, \cdot)$ represents a Normal distribution with standard deviation $\sigma$, $d_W$ is the expected value of a light measurement on the white circle, and $d_B$ is the expected value of a light measurement on the black background. Clearly, this can be made more accurate by working with the point-spread function, however, our aim here is to tie the inference engine to the inquiry engine in real-time.

The nested sampling algorithm samples circles with centers uniformly distributed across the field, and radii uniformly distributed from 1cm to 15cm. The result is a set of weighted samples from which the mean and the variance of the circle parameters can be estimated. From this set of weighted samples, we obtain a set of 150 circles distributed according to the posterior probability.

# The Inquiry Engine

This set of 150 circles is then used to examine the space of all possible measurements. This space is the set of locations in the field where the instrument can measure the light intensity. Each one of these possible measurements is a candidate experiment, so that choosing a measurement location is equivalent to designing an experiment. We will show that the fact that these circles are representative of the posterior probability simplifies the necessary computations. However, first we revisit the theory behind Bayesian adaptive estimation (Loredo, 2003).

Consider a proposed experiment $E$, which corresponds to taking a measurement at position $(x_e, y_e)$. We do not know for certain what we will measure, nor do we know the parameter values of our circle, but we can write the probability of the measurement $d_e$ in terms of the joint probability of $d_e$ and $\mathbf{C}$ as

$$p(d_e | \mathbf{D},(x_e,y_e),I) \equiv \int d\mathbf{C}\, p(d_e,\mathbf{C} | \mathbf{D},(x_e,y_e),I). \tag{9}$$

Using the product rule, we can write

$$p(d_e | \mathbf{D},(x_e,y_e),I) \equiv \int d\mathbf{C}\, p(d_e | \mathbf{C},\mathbf{D},(x_e,y_e),I)\, p(\mathbf{C} | \mathbf{D},(x_e,y_e),I). \tag{10}$$

This can be simplified by observing that, if we knew the circle parameters $\mathbf{C}$, we would not need the data $\mathbf{D}$

$$p(d_e | \mathbf{D},(x_e,y_e),I) \equiv \int d\mathbf{C}\, p(d_e | \mathbf{C},(x_e,y_e),I)\, p(\mathbf{C} | \mathbf{D},(x_e,y_e),I). \tag{11}$$

Probability theory only takes us so far. In this problem, we wish to make a decision, and this requires us to maximize the expected utility according to an assigned utility function: $U(\text{outcome, action})$, so that

$$(\hat{x}_e, \hat{y}_e) \equiv \int dd_e\, p(d_e | \mathbf{D},(x_e,y_e),I)\, U(d_e,(x_e,y_e)), \tag{12}$$

where the location $(x_e, y_e)$ is indicative of the action and the measurement $d_e$ is the outcome. Here we use a utility function based on the information provided by the measurement, so that we will choose the measurement that provides the greatest expected gain in information. Of course, other utility functions could be used that depend on the time it takes for the measurement to be taken, the energy required, etc. Utility functions such as these will surely be important in a fully-functioning automated instrument. Using the Shannon information for our utility function we find

$$U(d_e,(x_e,y_e)) = \int d\mathbf{C}\, p(\mathbf{C} | d_e,\mathbf{D},(x_e,y_e),I)\, \log p(\mathbf{C} | d_e,\mathbf{D},(x_e,y_e),I). \tag{13}$$

By writing the joint entropy for $\mathbf{C}$ and $d_e$, and writing the integral two ways, one can show (Loredo, 2003) that the optimal experiment can be found by maximizing the entropy of the possible measurements

$$(\hat{x}_e, \hat{y}_e) \equiv \arg\max_{(x_e,y_e)} \left(-\int dd_e\, p(d_e | \mathbf{D},(x_e,y_e),I)\, \log p(d_e | \mathbf{D},(x_e,y_e),I)\right). \tag{14}$$

This entropy can be easily estimated using the ensemble of models sampled from the posterior. For each measurement position $(x_e, y_e)$, we sample from the likelihood function of each sampled model thereby obtaining a set of potential measurements. The entropy of this set is rapidly estimated by constructing a histogram and computing the entropy directly. To enable the robot to consider a variety of positions, at each step we consider a grid on the space of possible measurements and compute the entropy only at the grid points. The alignment of this grid is randomly jittered so that a greater variety of points can be considered during the course of the experiment. With the optimal measurement position identified, the MATLAB software requests this particular measurement from the robotic instrument. Once the measurement is collected, the inference is updated, and the process is repeated until the system has estimated the model parameters with the desired accuracy.

## RESULTS

At this point, we are still working on obtaining a fully-functioning Bluetooth connection between the laptop computer running MATLAB and the NXT Brick. While we have tested the system by manually transmitting the information between the laptop and NXT Brick via a USB connection, in this presentation, we have simulated the process entirely in MATLAB. The result we present here is typical and dramatically demonstrates that the number of measurements required by an intelligent instrument is much smaller than a similar scanning system.

Figure 2A shows the initial stages of the inference-inquiry procedure where the white area of the circle has not yet been located. For this reason, there are large regions of the measurement space that are potentially equally informative. These are indicated by the large regions of essentially equal entropy in Figure 2B.

After several iterations, the robot finds a white area belonging to the circle. The set of sampled models are now close to the true circle (Figure 2C). The entropy map (Figure 2D) shows that the optimal measurement locations are those that are in the region where the models do not agree. This procedure naturally selects a binary question that at each stage rules out half of the models, which results in an extremely rapid convergence dramatically reducing the number of necessary measurements.

## CONCLUSION

This work constitutes an initial investigation into designing an intelligent instrument, which not only makes inferences from data, but also decides which measurements to take based on what the instrument has learned. The approach we have employed here relies on Bayesian adaptive exploration, which selects a measurement based on maximizing the entropy of the possible measurements obtained by querying a set of models sampled from the posterior. The results of this initial investigation reduces nicely to viewing the inquiry process as selecting efficient binary questions, which is known to be optimal from an information-theoretic perspective. It should be noted that these binary questions are not hard-wired into the system.

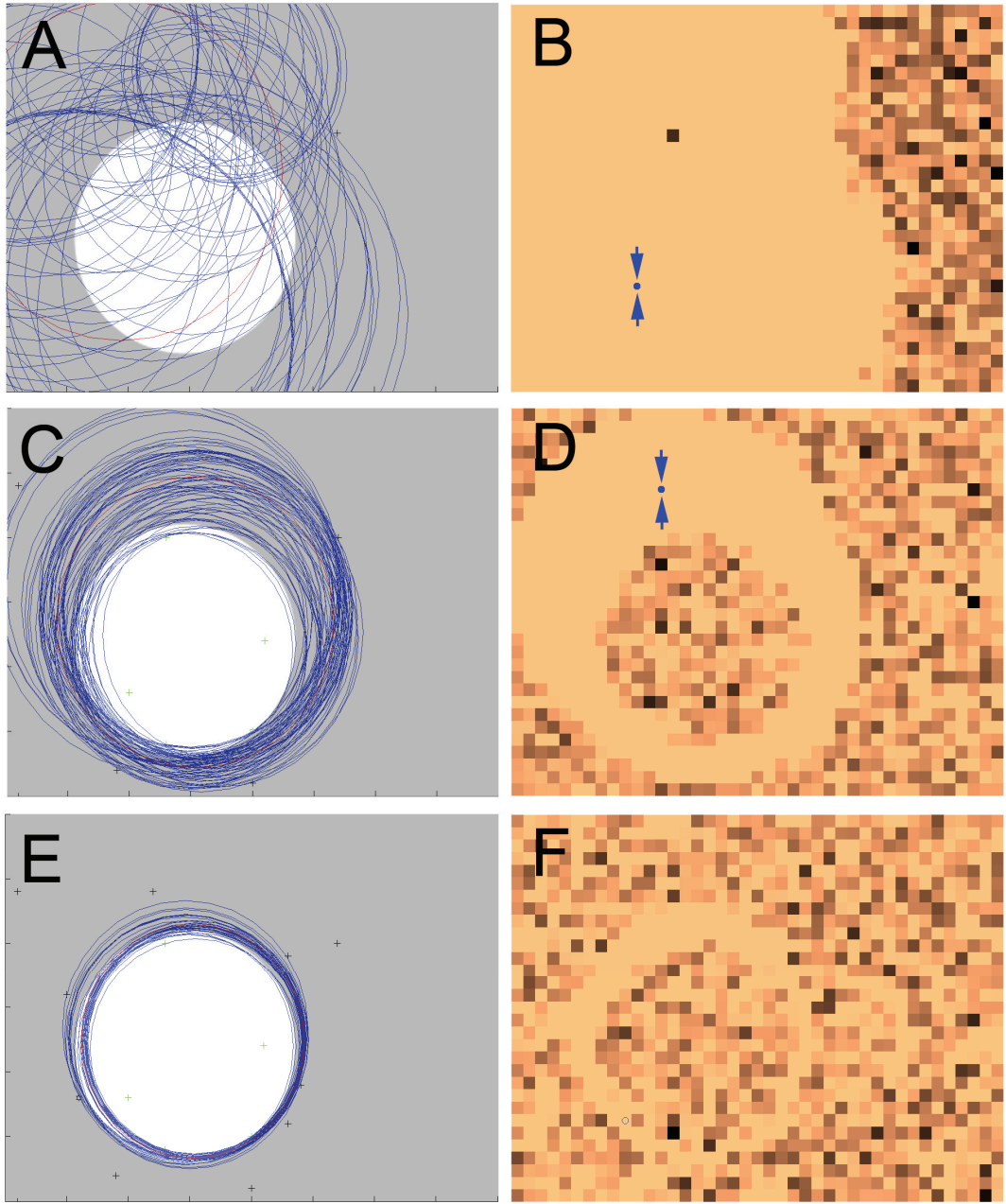

**FIGURE 2.** The panels on the left show the black playing field with the white circle. Overlaid on this are the set of 150 circles sampled from the posterior. Crosses indicate past measurement positions. The panels on the right show coarse entropy maps where the lighter shades indicate higher entropy. (A) One measurement has been taken (at the edge of the circles in upper right). This measurement has resulted in a set of hypothesized circles sampled from the posterior. (B) Much of the field is still unexplored indicated by the vast region of high entropy. The optimal measurement location is indicated by the dot with two arrows. (C) After 10 measurements, the algorithm is getting close to a solution. (D) Note that the region of high entropy is the region covered by the sampled circles. The chosen location divides the models into two. It will rule out half of the models with an efficient binary question. (E) After 16 measurements, the solution is almost obtained. (F) The corresponding entropy map is now focused on measurements at the edge of the circle.

Instead, they result as a natural application of maximizing the entropy of the potential measurement values given the model, the previous data, and our prior information.

This maximum entropy approximation works as long as the noise level, described by the likelihood function, is independent of the sampling location (Loredo, 2003). This condition will not always hold, and must be considered in future efforts.

Related maximum entropy techniques are finding their way into robotics (Thrun et al., 2005) and promise to enable these automated systems to interact with their environments in an intelligent manner. By creating joint environment-system models, the act of calibration becomes another potential experiment. In such a system, the instrument can decide to interact with the environment via measurement or itself via calibration giving rise to an instrument that actively self-calibrates during an experiment. Such advances are only the beginning.